\documentclass[sigconf, nonacm]{acmart}

\AtBeginDocument{%
  }

\usepackage{graphicx}
\usepackage{booktabs}
\usepackage{comment}
\usepackage{multirow}
\usepackage[table]{xcolor}
\usepackage{pifont}       
\newcommand{\cmark}{\ding{51}} 
\newcommand{\xmark}{\ding{55}} 

\newcommand{\methodname}{OCCAM}

\begin{document}

\title{OCCAM: Open-set Causal Concept explAnation and Ontology induction for black-box vision Models}

\author{Chiara Maria Russo}
\email{chiara.russo1@phd.unict.it}
\orcid{0009-0006-4402-6159}
\affiliation{%
  \institution{University of Catania}
  \city{Catania}
  \country{Italy}
}

\author{Simone Carnemolla}
\email{simone.carnemolla@phd.unict.it}
\orcid{0009-0005-8367-3933}
\affiliation{%
  \institution{University of Catania}
  \city{Catania}
  \country{Italy}
}

\author{Simone Palazzo}
\email{simone.palazzo@unict.it}
\orcid{0000-0002-2441-0982}
\affiliation{%
  \institution{University of Catania}
  \city{Catania}
  \country{Italy}
}

\author{Daniela Giordano}
\email{daniela.giordano@unict.it}
\orcid{0000-0001-5135-1351}
\affiliation{%
  \institution{University of Catania}
  \city{Catania}
  \country{Italy}
}

\author{Concetto Spampinato}
\email{concetto.spampinato@unict.it}
\orcid{0000-0001-6653-2577}
\affiliation{%
  \institution{University of Catania}
  \city{Catania}
  \country{Italy}
}

\author{Matteo Pennisi}
\email{matteo.pennisi@unict.it}
\orcid{0000-0002-6721-4383}
\affiliation{%
  \institution{University of Catania}
  \city{Catania}
  \country{Italy}
}

\renewcommand{\shortauthors}{Russo et al.}

\begin{abstract}
Interpreting the decisions of deep image classifiers remains challenging, particularly in black-box settings where model internals are inaccessible. We introduce OCCAM, a framework for open-set causal concept explanation and ontology induction in vision models. OCCAM discovers visual concepts in an open-set manner, localizes them via text-guided segmentation, and performs object-level interventions by removing concepts to measure changes in class confidence, estimating each concept’s causal contribution.

Beyond local explanations, OCCAM aggregates interventional evidence across a dataset to induce a structured concept ontology that captures how classifiers globally organize visual concepts. Reasoning over this ontology reveals consistent dependencies between concepts, exposes latent causal relations, and uncovers systematic model biases. Experiments on Broden and ImageNet-S across multiple classifiers show that OCCAM improves explanation quality in open-set black-box settings while providing richer global insight than per-image attribution methods.
\end{abstract}

\maketitle

\section{Introduction}
Deep image classifiers now reach near-human performance on many benchmarks, yet their decision-making process remains opaque, especially when models are accessible only through input--output queries. This strict black-box regime is the norm for deployed systems and third-party APIs, where internal gradients, activations, and architectural details are unavailable. At the same time, real-world images contain entities that extend beyond any predefined label vocabulary, making open-set concept discovery essential for meaningful explanations. Together, these constraints demand explanation methods that are simultaneously \emph{black-box}, \emph{open-set}, and \emph{causal}.

Existing approaches struggle in this regime. Attribution-based methods ~\cite{selvaraju2017grad, simonyan2013deep, sundararajan2017axiomatic, srinivas2019full, zeiler2014visualizing, han2025reasoning, fong2017interpretable, petsiuk2018rise, wagner2019interpretable, ribeiro2016should, lundberg2017unified} rely on internal gradients and provide continuous relevance maps without explicit semantic structure. Concept-based methods ~\cite{ahn2024unified, oikarinen2023clipdissect, kuroki2025fam, crabbe2022concept, fel2023craft, bau2017network, fong2018net2vec, koh2020concept, parchami2025fact, carnemolla2025dexter, park2018multimodal, hendricks2018grounding, sammani2022nlx, asgari2024texplain} often depend on predefined vocabularies or curated concept banks, limiting their ability to explain novel or fine-grained entities. Moreover, most techniques operate locally, explaining individual predictions without offering a principled mechanism to derive consistent global insights. As a result, they fail to uncover higher-order conceptual dependencies or systematic biases embedded in the model.

In this work, we introduce \textbf{OCCAM} (\textbf{O}pen-set \textbf{C}ausal \textbf{C}oncept expl\textbf{A}nation and ontology induction for black-box vision \textbf{M}odels), a framework that reframes explanation as a process of structured interventional pruning. Given an input image, OCCAM leverages a Multimodal Large Language Model (MLLM) to propose open-set concepts, grounds them spatially, and performs object-level deletions. By measuring the resulting prediction shifts, OCCAM estimates each concept’s causal contribution under strict query-only access.
Moreover, OCCAM's contribution goes beyond local causal probing. By aggregating interventional measurements across images, it induces a structured conceptual ontology that reflects how the classifier globally composes, prioritizes, and relates concepts. This ontology is not externally imposed, but emerges directly from interventional evidence.

We demonstrate that reasoning over the induced ontology enables explanations that reveal stable conceptual dependencies, expose compositional patterns, and uncover latent causal relations between entities that may remain undetected when analyzing images independently. 
Empirically, OCCAM achieves moderate improvements over prior concept-based methods ~\cite{ahn2024unified, oikarinen2023clipdissect, kuroki2025fam} on Broden ~\cite{bau2017network} and ImageNet-S ~\cite{gao2022large} across multiple architectures, while operating strictly in open-set and black-box conditions. We also show that the framework extends naturally to vision–language classifiers and that ontology-driven reasoning provides qualitatively deeper and more faithful explanations of model behavior.

\noindent Our overall contributions are:
\begin{itemize}
    \item OCCAM, a black-box framework for \emph{open-set causal explanation} via structured object-level interventions, requiring only query access to the visual classifier.
    \item A principled procedure that aggregates concept-level causal effects across images to construct an emergent global conceptual ontology of the model, without relying on predefined vocabularies or internal representations.
    \item We demonstrate that reasoning over the induced ontology enables deeper explanations, uncovering stable conceptual dependencies, compositional patterns, and latent causal relations that may remain undetected by isolated per-image explanations.
    \item State-of-the-art performance on Broden and ImageNet-S under strict open-set and black-box conditions, showing both quantitative improvements and qualitatively richer model understanding.
    \item Robustness analysis across vision–language models, highlighting differences in concept reliance across architectures.
\end{itemize}

\section{Related work}
Understanding what visual classifiers rely on at prediction time remains a central challenge in explainable AI. Approaches differ along several axes: \textit{intrinsic} vs.\ \textit{post-hoc}; explanatory unit (pixels, patches, semantic concepts); access to internals (white- vs.\ black-box); alignment with natural language; and whether explanations are \textit{local} or \textit{global}. Our method, \textbf{\methodname{}}, operates strictly in a \textit{post-hoc}, black-box, and open-set regime. It extracts image-specific concepts with an MLLM, grounds them spatially, measures their interventional necessity via object-level ablation, and aggregates these effects to construct a structured ontology capturing global model dependencies.

\textbf{Attribution-based explainability.}
Gradient-based saliency methods ~\cite{simonyan2013deep, sundararajan2017axiomatic} compute pixel importance via back-propagation. Class Activation Mapping (CAM) ~\cite{selvaraju2017grad} highlights class-specific regions, while feature visualization ~\cite{zeiler2014visualizing, srinivas2019full} synthesizes inputs activating neurons. Perturbation-based methods ~\cite{fong2017interpretable, wagner2019interpretable, petsiuk2018rise} occlude regions and measure prediction changes in black-box settings. Model-agnostic surrogates such as LIME ~\cite{ribeiro2016should} and SHAP ~\cite{lundberg2017unified} approximate local decision boundaries through perturbations. While useful, these methods operate over spatial units rather than semantic concepts, produce saliency signals requiring manual interpretation, and do not recover structured conceptual dependencies across inputs.

\textbf{Concept-based model dissection.}
Concept-based methods explain predictions in terms of predefined semantic entities by analyzing internal representations. Closed-set approaches ~\cite{kim2017interpretability, crabbe2022concept, zhou2018interpretable, fel2023craft} quantify sensitivity to annotated concepts using representation alignment or directional derivatives. Neuron-level analyses such as Network Dissection ~\cite{bau2017network} and Net2Vec ~\cite{fong2018net2vec} map internal units to semantic masks. Vision-language alignment methods including CLIP-Dissect ~\cite{oikarinen2023clipdissect}, WWW ~\cite{ahn2024unified}, and CE-FAM ~\cite{kuroki2025fam} estimate concept contributions via internal attribution mechanisms. While more interpretable than pixel-level saliency, these approaches require white-box access, operate within fixed vocabularies, and measure representational sensitivity rather than interventional necessity. They also do not construct structured global representations of concept dependencies.

\textbf{Open-set concept explanations.}
Open-vocabulary methods extend beyond predefined concept sets. Inherently explainable models such as CBMs and B-Cosification ~\cite{koh2020concept, parchami2025fact} introduce concept supervision during training but are not applicable \textit{post-hoc}. Natural Language Explanations and textual optimization approaches ~\cite{park2018multimodal, hendricks2018grounding, sammani2023uni, asgari2024texplain, zablocki2024gift, carnemolla2025dexter, pennisi2025diffexplainer} generate open-vocabulary rationales but remain descriptive rather than causally validated. Hierarchical explanation frameworks such as LVX ~\cite{yang2024language} organize concepts into structured trees, yet relations are imposed top-down rather than derived from measured model behavior.

\methodname{} differs by performing semantically grounded input-level interventions in a fully black-box, \textit{post-hoc} setting and aggregating measured necessity patterns across samples to induce a structured ontology. This ontology captures concept--class and inter-concept dependencies emerging directly from the classifier’s behavior.

\textbf{Ontologies and semantic knowledge representation.}
An ontology formally specifies a shared conceptualization of a domain ~\cite{gruber1993translation},
representing \textit{classes}, \textit{instances}, and \textit{relations}.
Ontologies support structured querying and reasoning ~\cite{berners2023semantic}. In contrast to prior works relying on externally curated ontologies, \methodname{} constructs a causally validated ontology derived from grounded concepts and interventional evidence. By aggregating object-level causal contributions, it produces an interrogable representation capturing individual, inter-concept, inter-class, and potentially inter-classifier dependencies.

Recent works explore ontology-aware LLM reasoning, including OG-RAG~~\cite{sharma2025og}, RoE~~\cite{han2025reasoning}, LOGicalThought~~\cite{nananukul2025logicalthought}, human--LLM collaboration~~\cite{garcia2025ontology}, and extraction systems such as OntoGPT~~\cite{caufield2024structured}. While these approaches assume predefined knowledge structures, \methodname{} provides an automatically derived, behavior-grounded ontology that enables structured global interrogation of model biases, dependencies, and semantic decision patterns.
\section{Method}
\begin{figure*}[t]
    \centering
    \includegraphics[width=0.95\linewidth]{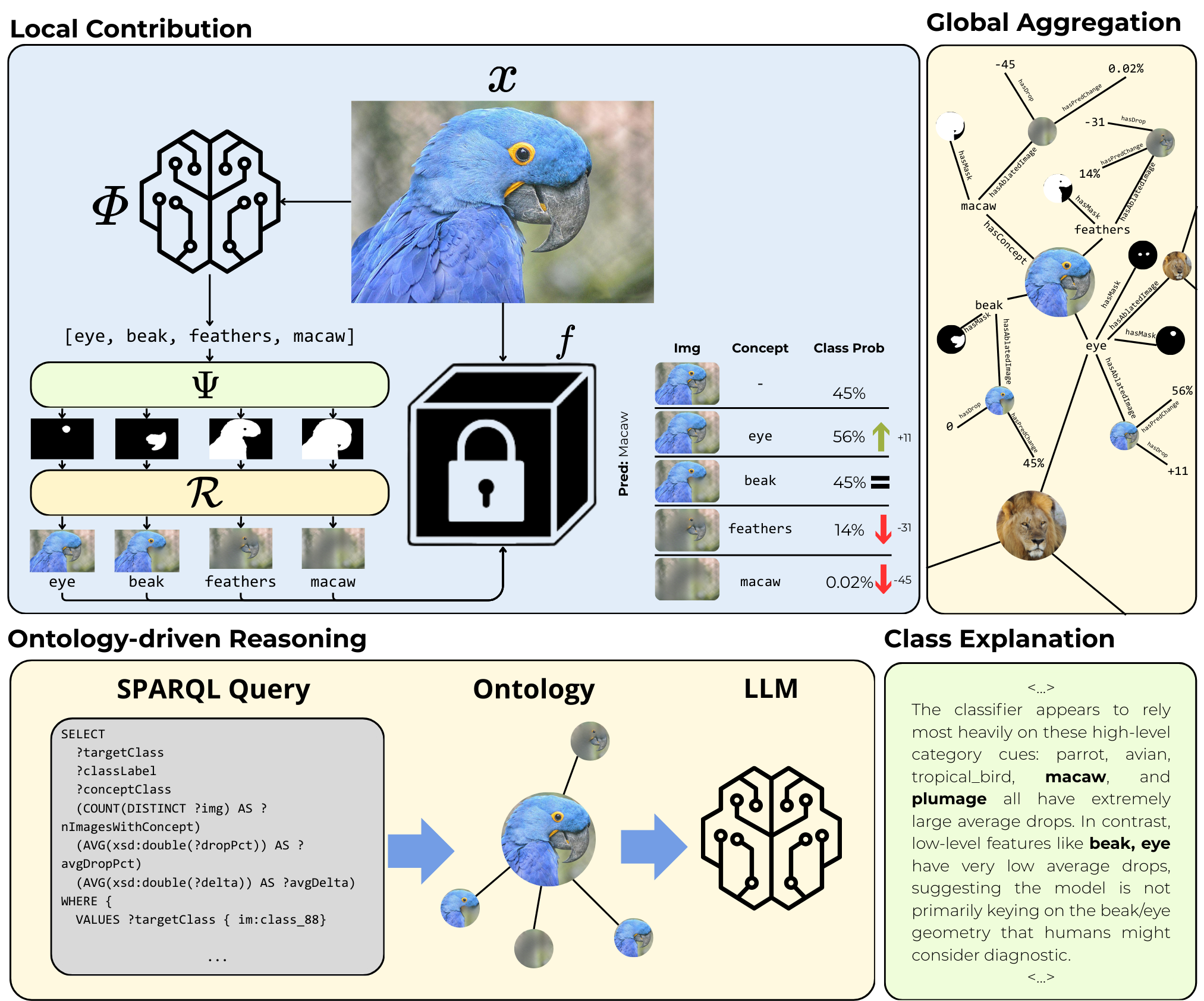}
    \caption{\textbf{OCCAM overview.} For an input image, open-set concepts are proposed, grounded spatially, and individually removed through input-level interventions. The induced prediction changes estimate causal contributions. Aggregating these interventional measurements across images enables ontology induction. 
    Structured querying enables to obtain global insights on model's decision-making process through ontology-driven LLM reasoning.} 
    \label{fig:method}
\end{figure*}

\textbf{OCCAM} is a concept-level interventional framework for explaining pre-trained image classifiers under strict black-box access and for inducing a structured conceptual ontology of their behavior. The key idea is to operationalize semantic interventions directly at the input level: by removing individual concepts and observing the resulting prediction change, we estimate their causal influence on the model's decision. By systematically aggregating these interventional effects across images, OCCAM constructs a global ontology that captures stable concept–class dependencies and higher-order conceptual relations, enabling reasoning beyond isolated per-image explanations. An overview of the proposed framework is illustrated in Fig.~\ref{fig:method}.

\subsection{Local interventional concept analysis}\label{local_concept_analysis}

Let $f: \mathcal{X} \rightarrow \mathbb{R}^{C}$ be a pretrained classifier mapping an image $x \in \mathcal{X}$ to a vector of $C$ class scores, and let $f_c(x)$ denote the $c$-th element of the vector. The predicted label is
\begin{equation}
\hat{y} = \arg\max_{c} f_c(x).
\end{equation}

We assume query-only access to $f$: given any image, we can observe output scores but cannot access internal weights, gradients, or intermediate representations. Consequently, explanations must be derived exclusively through controlled modifications of the input.

\paragraph{Open-set concept proposal.} We begin by identifying candidate semantic factors that may contribute to the prediction. Formally, we define a concept proposal operator
\begin{equation}
\Phi: \mathcal{X} \rightarrow \mathcal{P}\left(\mathcal{K}\right),
\end{equation}
where $\mathcal{P}(\mathcal{K})$ denotes the power set of the open-ended semantic universe $\mathcal{K}$ (e.g., the space of natural language descriptions). This operator maps an image to a finite set of instance-specific concepts from $\mathcal{K}$:
\begin{equation}
\Phi(x) = \{k_1, \dots, k_{N}\},
\end{equation}
with $N$ being input-dependent. Importantly, $\Phi$ operates in an \emph{open-set} regime: concepts are not drawn from a fixed vocabulary and may vary across images, allowing OCCAM to adapt explanations to the semantic content of each input.

\paragraph{Concept grounding.} Not all proposed concepts correspond to visually grounded entities. We therefore introduce a grounding operator
\begin{equation}
\Psi: \mathcal{X} \times \mathcal{K} \rightarrow \{0,1\}^{H \times W},
\end{equation}
which maps an image–concept pair $(x, k_i)$ to a binary segmentation mask
\begin{equation}
m_i = \Psi(x, k_i),
\end{equation}
representing the spatial support of concept $k_i$.
Concepts for which $\Psi$ fails to produce a valid mask are discarded\footnote{For notation simplicity and without loss of generality, we will continue addressing the set of remaining valid concepts as $\Phi(x) = \left\{k_1, \dots, k_N\right\}$.}.

\paragraph{Causal pruning.}
To estimate causal influence, we perform an input-level intervention that removes each grounded concept independently. We define an editing operator
\begin{equation}
\mathcal{R}: \mathcal{X} \times \{0,1\}^{H \times W} \rightarrow \mathcal{X},
\end{equation}
which produces an intervened image
\begin{equation}
\tilde{x}_i = \mathcal{R}(x, m_i).
\end{equation}

The image $\tilde{x}_i$ represents the input after removing the visual support of concept $k_i$, while preserving the remaining content (e.g., through inpainting). 

\paragraph{Causal contribution estimation.} For each grounded concept, we measure its causal contribution by evaluating the change in confidence for the originally predicted class:
\begin{equation}
s_{x,k_i} = f_{\hat{y}}(x) - f_{\hat{y}}(\tilde{x}_i),
\quad i \in \left\{1, \dots, N\right\}.
\end{equation}

The contribution score $s_{x,k_i}$ quantifies how much the classifier's evidence for $\hat{y}$ decreases when concept $k_i$ is removed. Larger values indicate that the concept provides stronger positive causal support for the decision. Concepts with negligible $s_{x,k_i}$ can be considered non-essential under this intervention, reflecting the structured pruning principle underlying OCCAM.

The complete interventional pipeline can therefore be summarized as:

\begin{equation}
x \xrightarrow{\Phi}
\begin{cases} 
k_1 \xrightarrow{\Psi} m_1 \xrightarrow{\mathcal{R}} \tilde{x}_1 \xrightarrow{f} s_{x,k_1} \\
\quad \vdots \\
k_N \xrightarrow{\Psi} m_N \xrightarrow{\mathcal{R}} \tilde{x}_N \xrightarrow{f} s_{x,k_N}
\end{cases}
\end{equation}

\subsection{Ontology construction and reasoning}
\label{sec:ontology}

While contribution scores provide per-image explanations, our goal is to transform them into a structured, dataset-level representation of model behavior.

We formalize the induced ontology as a knowledge graph, where nodes, i.e., ontology \emph{entities}, represent classes (\(c \in \{1,\dots,C\}\)), images ($x \in \mathcal{X})$, concepts (\(k \in \mathcal{K}\)), and causal evidence entities \(e_{x,k}\); the latter represent the evidence supporting the association between an image $x$ and a concept $k$, and their internal attributes include the contribution score $s_{x,k}$ and associated artifacts (e.g., concept grounding masks and edited images). Graph edges represent structural relations between these entities: for instance, \emph{concept associated to image}, \emph{image predicted as class}, \emph{evidence linked to image and concept}. 

Global concept–class and inter–concept dependencies are not imposed explicitly in the graph. Instead, they emerge from aggregating the stored interventional attributes across evidence instances. In this way, the ontology preserves the full local experimental trace while enabling dataset-level reasoning through structured queries.

Details of ontology schema, features and applications is provided in the supplementary material.
\label{method}
\section{Experiments}\label{results}
We evaluate \methodname{} along the three core components of the framework, addressing the following questions:
\begin{itemize}
    \item \emph{Grounding validity:} Are proposed open-set concepts spatially aligned with ground-truth regions?
    \item \emph{Causal validity:} Does input-level concept removal identify semantically meaningful and influential evidence for predictions?
    \item \emph{Global ontology-driven reasoning:} Does aggregating interventional evidence into a structured ontology enable coherent and informative class-level explanations?
\end{itemize}

\subsection{Experimental setup}

\noindent\textbf{Models.}

We evaluate \methodname{} on ResNet18, ResNet50 ~\cite{he2016deep}, and ViT-B/16 ~\cite{dosovitskiy2020image} --- two convolutional residual networks and a Vision Transformer --- as well as several CLIP-based models ~\cite{radford2021learning} (ViT-B/32, ViT-L/14, ViT-L/14-336) and SigLIP ~\cite{zhai2023sigmoid}, which are VLMs trained via contrastive image–text alignment. For a fair comparison with competitors, we used the same pre-trained versions as ~\cite{kuroki2025fam}.

\noindent\textbf{Datasets.}
We use semantic segmentation datasets commonly adopted for explainability evaluation. \textbf{Broden}~~\cite{bau2017network} contains 63k images and 1,197 concept labels spanning categories such as objects, parts, materials, and colors; following prior work ~\cite{kuroki2025fam}, we exclude scene and texture categories due to missing segmentation masks. \textbf{ImageNet-S}~~\cite{gao2022large} extends ImageNet-1k~~\cite{deng2009imagenet} with segmentation masks for 919 classes, enabling large-scale localization evaluation. We use the original validation splits, corresponding to 18,901 images for Broden and 12,419 for ImageNet-S.

\noindent\textbf{Implementation details}. The abstract operators $\Phi$, $\Psi$, and $\mathcal{R}$ are instantiated using pretrained foundation models. Open-set concepts are extracted with the MLLM \texttt{gemma3:27b} ~\cite{kamath2025gemma}. Concept grounding is performed using \texttt{SAM3} ~\cite{carion2025sam}, a text-conditioned segmentation model. Concept removal is implemented via the \texttt{LaMa} inpainting model ~\cite{suvorov2021lama}, ensuring coherent reconstruction of edited regions. The induced ontology is realized as an RDF/OWL knowledge graph. 

\noindent\textbf{Competitors.} We compare our method with State-of-the-Art approaches in cross-modal explainability. \textbf{CLIP-Dissect}~~\cite{oikarinen2023clipdissect} assigns semantic labels to individual neurons via similarity with CLIP text embeddings; \textbf{WWW}~~\cite{ahn2024unified} aligns intermediate features with CLIP and estimates concept importance using Shapley aggregation; \textbf{CE-FAM}~~\cite{kuroki2025fam} projects multi-layer features into CLIP space via training and gets concept localization maps from gradient-weighted activations.

Additional information on experimental setting, statistical significance analyses and system prompts are provided in the supplementary material.

\subsection{Metrics}\label{metrics}

For open-set concept grounding (Sect.~\ref{localization_eval}), we follow the evaluation protocol of CE-FAM~\cite{kuroki2025fam}. For causal pruning (Sect.~\ref{contribution}), we report metrics derived from causal contribution estimates produced by \methodname{}. These metrics quantify the effect of concept-level interventions introduced in Sect.~\ref{local_concept_analysis} and are stored as ontology properties. Importantly, all ontology-level quantities produced by \methodname{} are obtained in a fully black-box setting, without access to model internals; logit-based metrics are used exclusively for evaluation purposes (Sect.~\ref{contribution}).

\subsubsection{Causal pruning metrics}\label{metrics_contribution}

Let $f(x) \in \mathbb{R}^{C}$ denote the classifier output scores. The predicted class on the original image is
\[
\hat{y} = \arg\max_c f_c(x),
\]
with corresponding score
\[
s = f_{\hat{y}}(x),
\qquad
s_i = f_{\hat{y}}(\tilde{x}_i).
\]

Ontology-level metrics include the post-intervention score $s_i$, the confidence change $s_{x,k_i}$ (as defined in Sect. \ref{local_concept_analysis}), and the concept \textbf{Mask Area (MA)}. 

The relative area of the segmentation mask is
\[
\text{MA}_i =
100 \cdot
\frac{1}{HW}
\sum_{u=1}^{H}
\sum_{v=1}^{W}
\mathbf{1}\{m_i(u,v) > 0\}
\]
which measures the percentage of the image occupied by the concept mask; $m_i(u, v)$ denotes a single pixel in the mask, and $\mathbf{1}(\cdot)$ is the indicator function.

These quantities are computed solely from classifier outputs and segmentation masks.

For evaluation, we compute additional metrics that quantify the effect of concept removal on model confidence and log-odds. These include the \textbf{Confidence Drop Percentage (CDP)}:

\[
\text{CDP}_i =
100 \cdot \frac{\max(0, s - s_i)}{s + \varepsilon},
\]
the log-odds change \textbf{Logit Delta (LD)}:
\[
\text{LD}_i = \ell(s_i) - \ell(s),
\]

and the relative log-odds drop $\text{pct\_logit\_drop}_i$.

From these per-concept quantities, we derive aggregated metrics used for evaluation reported in (Table~\ref{tab:concept_contribution}):

\begin{enumerate}
    \item \textbf{Average Drop Percentage (ADP):} mean confidence drop percentage across all removed concepts in an image,
    \[
    \mathrm{ADP} =
    \frac{1}{N}\sum_{i=1}^{N} \text{CDP}_i.
    \]

    \item \textbf{Maximum Drop Percentage (MDP):} largest confidence drop percentage among all removed concepts,
    \[
    \mathrm{MDP} =
    \max_{i \in \{1,\dots,N\}} \text{CDP}_i.
    \]

    \item \textbf{Maximum Absolute Logit Drop (MAD):} maximum absolute change in log-odds induced by any single concept removal,
    \[
    \mathrm{MAD} =
    \max_{i \in \{1,\dots,N\}} \left| \text{LD}_i \right|.
    \]
\end{enumerate}

\subsubsection{Open-set concept grounding metrics}

We evaluate concept localization following the protocol of~\cite{kuroki2025fam}, using complementary metrics (EPG, NRA, and Hit Rate) to compare predicted activation maps with ground-truth segmentation masks. Let
\[
A \in \mathbb{R}^{H \times W}, \quad
M \in \{0,1\}^{H \times W}
\]
denote the predicted activation map and the binary mask, respectively.

\paragraph{Energy-based Pointing Game (EPG)} EPG measures the fraction of activation energy falling within the ground-truth region. The activation map is normalized to $[0,1]$ as
\[
\tilde{A} = \frac{A - \min(A)}{\max(A) - \min(A)},
\]
with score set to $0$ if $\max(A)=\min(A)$. The metric is defined as
\[
\mathrm{EPG}(A,M) =
\frac{\sum_i \tilde{A}_i M_i}{\sum_i \tilde{A}_i},
\]
and is $0$ if the denominator vanishes.

\paragraph{Normalized Region Accuracy (NRA)} NRA evaluates alignment between activation maps and ground-truth masks via thresholded regions. For a map $R$, we define regions $(R \ge T_n)$ selecting the top $n\%$ activations and compute:
\[
\mathrm{IoU}_n =
\frac{|(R \ge T_n) \cap M|}{|(R \ge T_n) \cup M|}.
\]
Varying $T_n$ yields an IoU curve whose area is normalized using random and ideal baselines:
\[
\mathrm{NRA} =
\frac{\mathrm{AUC} - \mathrm{AUC}_{\text{low}}}
{\mathrm{AUC}_{\text{high}} - \mathrm{AUC}_{\text{low}}}.
\]

\paragraph{Hit Rate} Hit Rate is defined as the fraction of concept–image pairs with $\mathrm{NRA} > 0.5$, indicating successful localization:
\[
\mathrm{Hit} \ \mathrm{Rate} =
\frac{1}{N}\sum_{j=1}^{N} \mathbf{1}\{\mathrm{NRA}_j > 0.5\}.
\]

\subsection{Causal pruning results}\label{contribution}

First, we evaluate whether removing a grounded concept induces a decrease  in the classifier’s confidence for the originally predicted class. 
This defines a causal pruning task: if a concept provides necessary evidence for a decision, eliminating its spatial support should reduce the predicted class logit.

Since our evaluation protocol requires removing a well-defined spatial support for each concept, and for CLIP-Dissect and WWW the concept-neuron relationship is inherently many-to-many, we restrict the comparison to CE-FAM, which explicitly produces per-image concept-level spatial grounding. To enable a direct comparison, both CE-FAM and \methodname{} are evaluated under the same input-level intervention procedure.
For each attributed concept, its grounded spatial region is removed, and the change in the predicted-class logit is measured.

Experiments are conducted on Broden, which provides multi-granular semantic annotations. 
To avoid degenerate cases, we exclude masks covering $\geq$99\% of the image. The resulting paired evaluation set contains 15,380 images per architecture.

To quantify causal pruning, we measure the drop in the predicted-class logit after concept removal.  
For each model, we report: 1) \textbf{ADP}, capturing the typical causal influence of attributed concepts across images; 2) \textbf{MDP}, reflecting whether the method identifies the most decision-critical concept per image; and 3) \textbf{MAD}, measuring the strength of the strongest intervention in logit space (Sect. \ref{metrics_contribution}).

\begin{table}[!t]
\centering
\caption{Concept contribution results on the Broden dataset.
Higher values indicate stronger causal contribution of identified concepts.}
\label{tab:concept_contribution}
\renewcommand{\arraystretch}{1.4}
\setlength{\tabcolsep}{8pt}

\begin{tabular}{l l c c c}
\toprule
 &  & ADP & MDP & MAD \\
\midrule
\multirow{2}{*}{ResNet50} 
  & CE-FAM~\cite{kuroki2025fam} & 44.18 & 84.14 & 3.37 \\
  & \textbf{OCCAM}   & \textbf{47.19} & \textbf{93.89} & \textbf{4.57} \\
\midrule
\multirow{2}{*}{ResNet18} 
  & CE-FAM~\cite{kuroki2025fam} & 49.12 & 87.78 & 4.80 \\
  & \textbf{OCCAM}   & \textbf{51.82} & \textbf{95.93} & \textbf{6.56} \\
\midrule
\multirow{2}{*}{ViT-B/16} 
  & CE-FAM~\cite{kuroki2025fam} & 46.20 & 87.04 & 5.08 \\
  & \textbf{OCCAM}   & \textbf{49.66} & \textbf{96.24} & \textbf{6.76} \\
\bottomrule
\end{tabular}

\end{table}

Table~\ref{tab:concept_contribution} summarizes results for ResNet50,  ResNet18, and ViT-B/16. \methodname{} produces valid grounded concepts for 99.6\% of images, compared to 82.0\% for CE-FAM, indicating greater robustness in  generating evaluable concept regions.
On valid samples, \methodname{} consistently induces stronger confidence reductions, demonstrating that object-level interventional grounding effectively identifies and prunes decision-critical evidence, yielding stronger and more reliable estimates of 
concept-level causal contribution.

Furthermore, to ensure that observed logit reductions are not caused by visual  artifacts introduced by inpainting, we measure the Fréchet Inception Distance (FID) between original and edited images. We obtain FID scores of 0.76 on Broden and 1.20 on ImageNet-S, indicating negligible distributional shift. Thus, the observed confidence drops reflect genuine causal effects 
of removing semantically grounded regions rather than perceptual degradation.

\subsection{Open-set concept grounding results}\label{localization_eval}

\begin{table}[!t]
\centering
\setlength{\tabcolsep}{4pt}
\caption{Localization results on the ImageNet-S validation split. }

\begin{tabular}{llccccc}
\toprule
~~~~~  & & BB & OS & EPG & NRA & Hit Rate \\
\midrule

\rowcolors{1}{gray!12}{white}

\multirow{4}{*}{\rotatebox[origin=c]{90}{ResNet50}}
& CLIP-Dissect~\cite{oikarinen2023clipdissect} & \xmark & \xmark & 0.61 & 0.59 & 0.66 \\
& WWW~\cite{ahn2024unified}         & \xmark & \xmark & 0.63 & 0.60 & 0.67 \\
& CE-FAM~\cite{kuroki2025fam}      & \xmark & \xmark & 0.62 & \textbf{0.61} & \textbf{0.74} \\
& \textbf{OCCAM }         & \cmark & \cmark & \textbf{0.65} & 0.56 & 0.68 \\

\midrule

\multirow{4}{*}{\rotatebox[origin=c]{90}{ResNet18}}
& CLIP-Dissect~\cite{oikarinen2023clipdissect} & \xmark & \xmark & 0.50 & \textbf{0.56} & 0.26 \\
& WWW~\cite{ahn2024unified}         & \xmark & \xmark & 0.50 & 0.53 & 0.14 \\
& CE-FAM~\cite{kuroki2025fam}      & \xmark & \xmark & 0.53 & 0.49 & 0.53 \\
& \textbf{OCCAM }         & \cmark & \cmark & \textbf{0.65} & \textbf{0.56} & \textbf{0.68} \\

\midrule

\multirow{4}{*}{\rotatebox[origin=c]{90}{ViT-B/16}}
& CLIP-Dissect~\cite{oikarinen2023clipdissect} & \xmark & \xmark & 0.24 & 0.17 & 0.05 \\
& WWW~\cite{ahn2024unified}         & \xmark & \xmark & 0.44 & 0.45 & 0.04 \\
& CE-FAM~\cite{kuroki2025fam}      & \xmark & \xmark & \textbf{0.67} & \textbf{0.67} & \textbf{0.78} \\
& \textbf{OCCAM }         & \cmark & \cmark & 0.64 & 0.55 & 0.66 \\

\bottomrule
\end{tabular}

\label{tab:imagenet_results}
\end{table}

Then, we evaluate whether OCCAM’s concept proposal ($\Phi$) and grounding ($\Psi$) operators produce consistent and spatially aligned concept regions that correspond to ground-truth semantic masks. 
In this scenario, we compare OCCAM against all three concept-based methods that associate semantic entities with internal representations under a predefined vocabulary: CLIP-Dissect~~\cite{oikarinen2023clipdissect}, WWW~~\cite{ahn2024unified} and CE-FAM~~\cite{kuroki2025fam}.\\
Note that all baselines require access to internal activations and operate in a closed-set vocabulary, whereas OCCAM works under strict black-box access and open-set concept discovery.

For localization-based evaluation, we adopt the unified protocol of CE-FAM~\cite{kuroki2025fam}, which compares continuous concept activation maps against ground-truth segmentation masks. CE-FAM derives concept-specific localization maps from internal feature activations. In contrast, OCCAM’s grounding operator $\Psi(x,k)$ produces a binary segmentation mask indicating the spatial support of concept $k$. 
To ensure comparability with the CE-FAM protocol and take the classifier's behavior into account, we convert this binary mask into a continuous concept-specific activation map by modulating it with the class-conditioned Grad-CAM map of the image. We would like to underline that, for these results, Grad-CAM maps are computed solely for evaluation and comparison with other methods. The proposed method itself does not require the computation of activation maps and does not rely on access to model weights.

On the binary mask, we then measure spatial agreement with ground-truth through three complementary metrics defined in~~\cite{kuroki2025fam}: 1) \textbf{EPG}, which measures the fraction of activation mass falling inside the ground-truth region; 2) \textbf{NRA}, which evaluates how well the highest-activated pixels recover the ground-truth mask under progressive inclusion, assessing ranking quality; and 3) \textbf{Hit Rate}, the fraction of concept–image pairs with $\mathrm{NRA}>0.5$, indicating successful localization. Details on metrics are provided in Sect. \ref{metrics}.

Since OCCAM proposes unconstrained natural-language concepts, we perform semantic alignment between predicted concepts and ground-truth labels using cosine similarity between Sentence-transformers embeddings~\cite{reimers-2019-sentence-bert}.  
We exclude the Broden dataset from this evaluation to avoid mismatch in many-to-many concept association, while for ImageNet-S, which provides a single mask per image, we select the top-1 predicted concept based on similarity to the ground-truth label.
Evaluation is performed only on aligned concept–image pairs.

Table~\ref{tab:imagenet_results}, which is an extension of CE-FAM's~\cite{kuroki2025fam} published results, presents results on ImageNet-S, showing that while operating under black-box and open-set conditions, OCCAM exhibits competitive or superior performance w.r.t. white-box and closed-set competitors.

Overall, across architectures and datasets, OCCAM matches or surpasses prior methods while imposing stricter constraints (no internal access and open-set discovery), indicating that object-level interventional grounding produces spatially aligned and semantically meaningful concept regions.

\subsection{Ontology-driven reasoning evaluation}\label{llm-judges}
\label{ontology}

We evaluate whether aggregating interventional evidence into a structured ontology improves global reasoning about the classifier’s decision mechanism. 
Specifically, we assess whether ontology-grounded knowledge enables more coherent and faithful class-level explanations compared to unstructured or flat structured representations.

We consider three reasoning settings: 

1) \textbf{LLM (unstructured)}: the model receives OCCAM’s interventional results as unstructured textual summaries describing concept frequencies and average causal effects; 

2) \textbf{LLM + JSON}: the same interventional statistics are provided in a flat structured JSON format; 

3) \textbf{LLM + ontology}: the identical evidence is organized within the induced ontology, allowing the LLM to retrieve aggregated concept–class dependencies through structured graph queries.

Each reasoning system generates a global report describing the decision mechanism of the classifier for a target class, given class name, assigned concepts and relative evidence scores. An excerpt from an actual generated report is shown in Fig.~\ref{fig:method}; additional examples are provided in the supplementary material.

The evaluation focuses on a subset of five target classes randomly selected from ImageNet-S. To avoid circular evaluation bias, reports are assessed by four different LLMs from those used for generation. In addition, 30 human evaluators provide independent ratings\footnote{Human evaluators were recruited through Amazon Mechanical Turk and are not domain experts.}. These human evaluations should not be interpreted as a formal user study, but rather as a control group used to verify the grounding of the judgments produced by the LLM evaluators.

Judges rate each report on a 1–5 scale based on the grounding of the explanation with respect to a set of images. 

Table~\ref{tab:eval_scores} shows that 
merely structuring knowledge in JSON does not improve reasoning quality compared to unstructured summaries.
In contrast, ontology-grounded reasoning consistently achieves the highest scores across all judge groups. 

Notably, all settings achieve consistently high scores (above 4 on average), indicating that the underlying causal pruning mechanism already produces semantically meaningful and well-grounded evidence for global explanation. 
Ontology-driven reasoning further improves upon this strong baseline, with consistent gains across both LLM judges and human evaluators. The agreement across independent AI and human judges suggests that while causal interventions provide reliable explanatory signals, organizing them within a formally structured ontology enhances coherence, faithfulness, and interpretability at the global level.

\paragraph{Reasoning and consistency checking.}
Unlike most structured data formats, an ontology comes with a formal semantics and can therefore be \emph{validated} with automated reasoning. An ontological reasoner can infer implicit facts entailed by the schema (e.g., subclass membership) and check whether the asserted triples are logically consistent with declared constraints (e.g., domain/range and class disjointness, when specified). Although our analyses mainly use explicitly stored interventional evidence, we run the HermiT reasoner~\cite{glimm2014hermit} to perform \textbf{consistency checking} of the exported graph. This verifies that the ontology respects the intended typing---for example, that intervention nodes only connect to images and concept instances, and that visual artifacts remain typed as images. Consistency checking becomes especially valuable when the ontology is extended or when graphs from multiple classifiers and datasets are merged into a shared knowledge base.

\begin{table}[t]
\centering
\caption{Global explanation quality scores (1--5) assigned by independent LLM and human judges under three knowledge settings: unstructured summaries (LLM), structured JSON, and ontology-grounded reasoning. Higher is better.}\label{tab:eval_scores}

\setlength{\tabcolsep}{6pt} 
\begin{tabular}{lccc}
\toprule
Judge & LLM & LLM + JSON & LLM + Ontology \\
\midrule
ChatGPT 5.2 & 4.60 & 4.20 & \textbf{4.80} \\
Gemini 3    & 4.00 & 4.20 & \textbf{4.80} \\
Qwen 235B   & 4.00 & 4.00 & \textbf{5.00} \\
Gemma 27B   & 4.40 & 4.40 & \textbf{4.60} \\
Human       & 4.55 & 4.33 & \textbf{4.70} \\
\midrule
Avg.        & 4.31 & 4.23 & \textbf{4.78} \\
\bottomrule
\end{tabular}
\end{table}

\subsection{Extending OCCAM to Multi-Modal Classifiers}
\label{clip}

\begin{table}[t]
\centering
\setlength{\tabcolsep}{6pt}
\caption{Accuracy comparison across CLIP variants under different numbers of removed concepts ($k$).}
\label{tab:clip_comparison}
\begin{tabular}{lcccc}
\toprule
 & \multicolumn{4}{c}{\textbf{Accuracy}} \\
\cmidrule(lr){2-5}
 & $k=0$ & $k=1$ & $k=2$ & $k=3$ \\
\midrule
CLIP ViT-B/16     & 0.69 & 0.23 & 0.11 & 0.07 \\
CLIP ViT-B/32     & 0.64 & 0.22 & 0.11 & 0.08 \\
CLIP ViT-L/14     & 0.76 & 0.26 & 0.13 & 0.08 \\
CLIP ViT-L/14-336 & 0.77 & 0.27 & 0.13 & 0.08 \\
SigLIP            & \textbf{0.78} & \textbf{0.29} & \textbf{0.17} & \textbf{0.14} \\
\bottomrule
\end{tabular}

\end{table}

\begin{figure*}[t!]
    \centering
    \includegraphics[width=.8\linewidth]{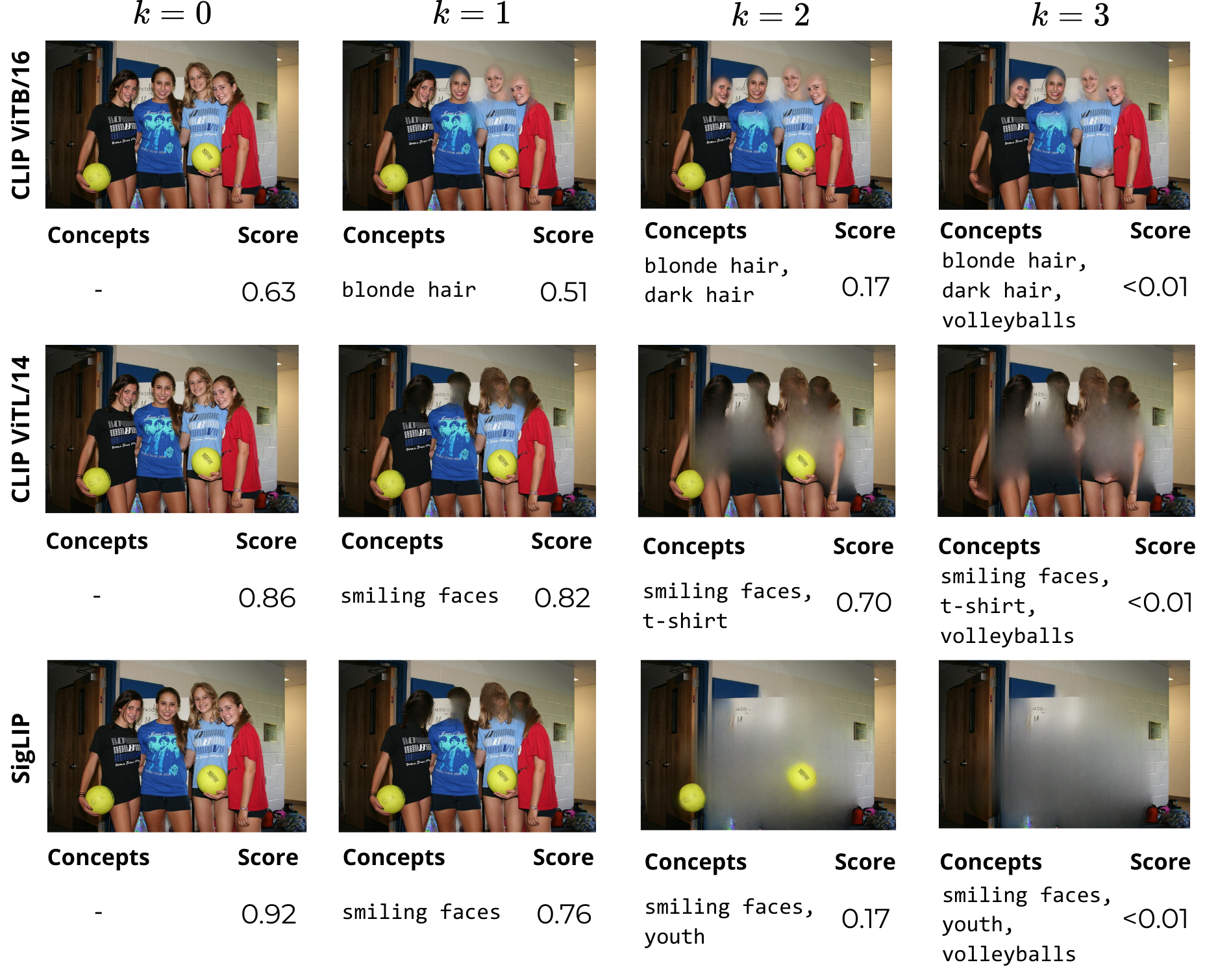}
\caption{Progressive causal pruning on the \texttt{volleyball} class for different multimodal classifiers. 
For each model, the top-3 influential concepts identified by OCCAM are removed sequentially ($k=1,2,3$). 
Prediction confidence decreases as explanatory factors are eliminated, revealing how different architectures distribute reliance over object and contextual cues.}
    \label{fig:clip_results}
\end{figure*}

Finally, we evaluate whether OCCAM’s input-level interventional framework extends naturally to multi-modal image--text classifiers, where predictions are obtained via zero-shot image--text similarity scores.

We consider several CLIP variants (ViT-B/16, ViT-B/32, ViT-L/14, ViT-L/14-336) and SigLIP. 
All models expose a black-box scoring function and therefore remain compatible with OCCAM’s concept removal strategy.

For each classifier, we apply OCCAM to identify influential open-set concepts and compute a size-normalized importance score to avoid favoring concepts that simply occlude large image regions. 
Concept importance is aggregated across images to obtain a classifier-specific ranking.
We then select the top-3 most influential concepts per classifier and progressively remove them from each image: i) remove the least influential among the top-3 ($k=1$), ii) remove the two least influential ($k=2$),
iii) remove all three simultaneously ($k=3$). Zero-shot classification accuracy is measured at each stage 
($k=0$ corresponds to original images).

Table~\ref{tab:clip_comparison} reports accuracy under increasing numbers of removed concepts. 
Across all models, accuracy decreases monotonically as $k$ increases, indicating that OCCAM successfully identifies regions that causally support the prediction.

Notably, SigLIP remains consistently more robust than CLIP variants under the same intervention budget, preserving non-trivial performance even after removing the top-3 influential concepts. 
This suggests that SigLIP’s decisions are less concentrated on a small set of high-impact regions and may rely on more distributed evidence.

Fig.~\ref{fig:clip_results} provides a qualitative example for the \texttt{volleyball} class, illustrating how different multi-modal classifiers surface distinct influential concepts and how prediction confidence degrades as explanatory factors are removed. For instance, all classifiers seem to share the importance of \textit{hair} (either blonde or dark), CLIP ViT-L/14 and SigLIP prove to be sensitive to \textit{smiling faces}, and SigLIP specifically identifies \textit{youth} as one of the core concepts for predicting the \texttt{volleyball} class.   

These results demonstrate that OCCAM’s open-set concept grounding and causal pruning generalize beyond purely visual classifiers. The framework remains applicable to image--text models without requiring access to internal representations, confirming that input-level interventions provide a unified explanation mechanism across unimodal and multimodal architectures.
\section{Conclusion}
We introduced OCCAM, a strictly black-box framework for open-set causal concept explanation and ontology induction in vision models. By performing semantically grounded input-level interventions, OCCAM estimates concept necessity without accessing internal representations, enabling explanation under realistic query-only constraints. 

Beyond per-image attribution, we show that systematically aggregating interventional evidence across large-scale datasets induces a structured conceptual ontology that captures stable concept–class dependencies and higher-order inter-concept relations. This ontology is not externally imposed but emerges directly from measured causal effects, preserving the full experimental trace while enabling structured global reasoning.

Empirical results on Broden and ImageNet-S demonstrate that OCCAM achieves strong spatial grounding under open-set conditions and produces more impactful causal pruning than prior concept-based methods. Furthermore, ontology-driven aggregation consistently improves global explanation quality across both LLM-based and human evaluations, indicating that formally structured knowledge enhances coherence and faithfulness beyond flat structured summaries.

Importantly, we demonstrated that OCCAM naturally extends to multimodal classifiers such as CLIP and SigLIP, confirming that input-level concept interventions provide a unified explanation mechanism across unimodal and image–text architectures.

Overall, OCCAM reframes model interpretation as structured interventional pruning: by identifying and removing causally essential concepts, it constructs a compact yet expressive representation of how classifiers organize visual evidence. We hope this perspective encourages future work on behavior-grounded knowledge induction and structured reasoning for explainable AI.

\bibliographystyle{ACM-Reference-Format}
\bibliography{main}

\end{document}